\documentclass[10pt,twocolumn,letterpaper]{article}

\usepackage{cvpr}              
\usepackage{cuted}
\usepackage{float}
\usepackage{algorithm}
\usepackage{algorithmic}
\definecolor{cvprblue}{rgb}{0.21,0.49,0.74}
\usepackage[pagebackref,breaklinks,colorlinks,allcolors=cvprblue]{hyperref}

\def\paperID{4797} 
\def\confName{CVPR}
\def\confYear{2025}

\title{TransPixeler: Advancing Text-to-Video Generation with Transparency}

\author{
Luozhou Wang\textsuperscript{$1$}\thanks{This work was done during an internship at Adobe Research.} \quad
Yijun Li\textsuperscript{$3$}\thanks{Project Leader}\quad
Zhifei Chen\textsuperscript{$1$} \quad
Jui-Hsien Wang\textsuperscript{$3$} \quad
Zhifei Zhang\textsuperscript{$3$} \quad
He Zhang\textsuperscript{$3$} \quad \\
Zhe Lin\textsuperscript{$3$} \quad
Yingcong Chen\textsuperscript{$1,2$}\thanks{Corresponding author} \quad
\\
$^1$ HKUST(GZ).\quad 
$^2$ HKUST.\quad 
$^3$ Adobe Research.
}

\begin{document}
\maketitle
\begin{strip}
    \centering
    \vspace{-5em}
    \centering
    \includegraphics[width=\textwidth]{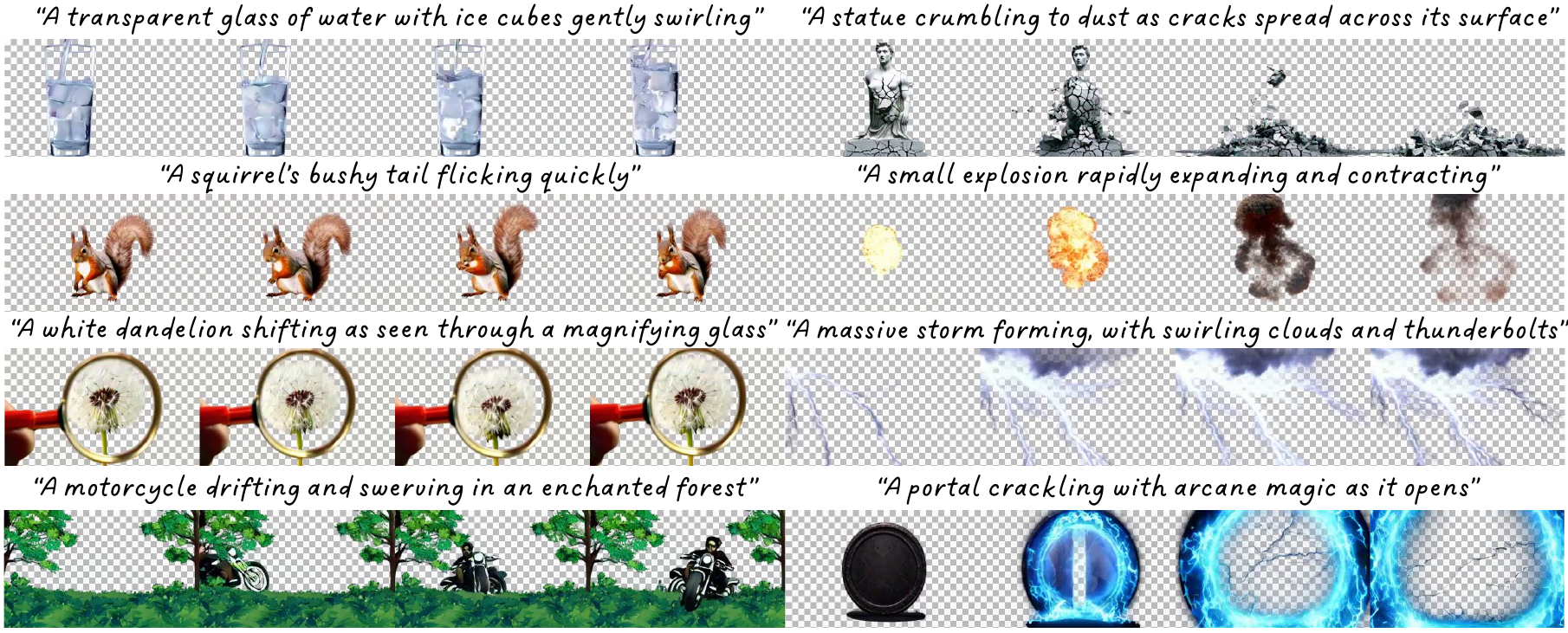}
    \vspace{-2em}
    \captionof{figure}{\textbf{RGBA Video Generation with TransPixeler.} By introducing LoRA layers into DiT-based text-to-video model with a novel alpha channel adaptive attention mechanism, our method enables RGBA video generation from text while preserving Text-to-Video quality. 
    }
    \label{fig:teaser}
\end{strip}

\begin{abstract}
Text-to-video generative models have made significant strides, enabling diverse applications in entertainment, advertising, and education. However, generating RGBA video, which includes alpha channels for transparency, remains a challenge due to limited datasets and the difficulty of adapting existing models. Alpha channels are crucial for visual effects (VFX), allowing transparent elements like smoke and reflections to blend seamlessly into scenes.
We introduce TransPixeler, a method to extend pretrained video models for RGBA generation while retaining the original RGB capabilities. TransPixeler leverages a diffusion transformer (DiT) architecture, incorporating alpha-specific tokens and using LoRA-based fine-tuning to jointly generate RGB and alpha channels with high consistency. By optimizing attention mechanisms, TransPixeler preserves the strengths of the original RGB model and achieves strong alignment between RGB and alpha channels despite limited training data.
Our approach effectively generates diverse and consistent RGBA videos, advancing the possibilities for VFX and interactive content creation. The code is available at \href{https://wileewang.github.io/TransPixeler/}{https://wileewang.github.io/TransPixeler/}. 
\end{abstract}    
\section{Introduction}
\label{sec:intro}
Text-to-Video generative models have quickly advanced, achieving impressive results~\cite{he2022latent, chen2023videocrafter1, guo2023animatediff, wang2023modelscope, wang2024videocomposer, zhang2023show, yang2024cogvideox, opensora, opensoraplan}. This progress has enabled various applications, such as video editing~\cite{geyer2023tokenflow, yang2023rerender, qi2023fatezero, liu2024video, wu2023tune, chen2023control}, image animation~\cite{blattmann2023stable, niu2024mofa, guo2024liveportrait, guo2023i2v}, and motion customization~\cite{he2024cameractrl, wang2024motionctrl, ling2024motionclone, jeong2024dreammotion, ma2023trailblazer, yin2023dragnuwa, wang2024motion}. Diffusion Transformers (DiT) enhance these models by using self-attention to capture long-range dependencies~\cite{sora2024, opensora, opensoraplan, yang2024cogvideox}. These models are now widely used in entertainment, advertising, and education, meeting the demand for customizable, dynamic content. 
Notably, Text-to-RGBA (A denotes Alpha channel) video generation is invaluable for VFX and creative industries. The inclusion of an alpha channel in RGBA formats allows for transparent effects, enabling seamless blending of elements like smoke and reflections (see Fig.~\ref{fig:teaser}). This transparency creates realistic visuals that can integrate smoothly into scenes without modifying the background. Such flexibility is crucial in gaming, virtual reality (VR), and augmented reality (AR), where dynamic and interactive content is in high demand.

\begin{figure}[t]
    \centering
    \includegraphics[width=1.0\linewidth]{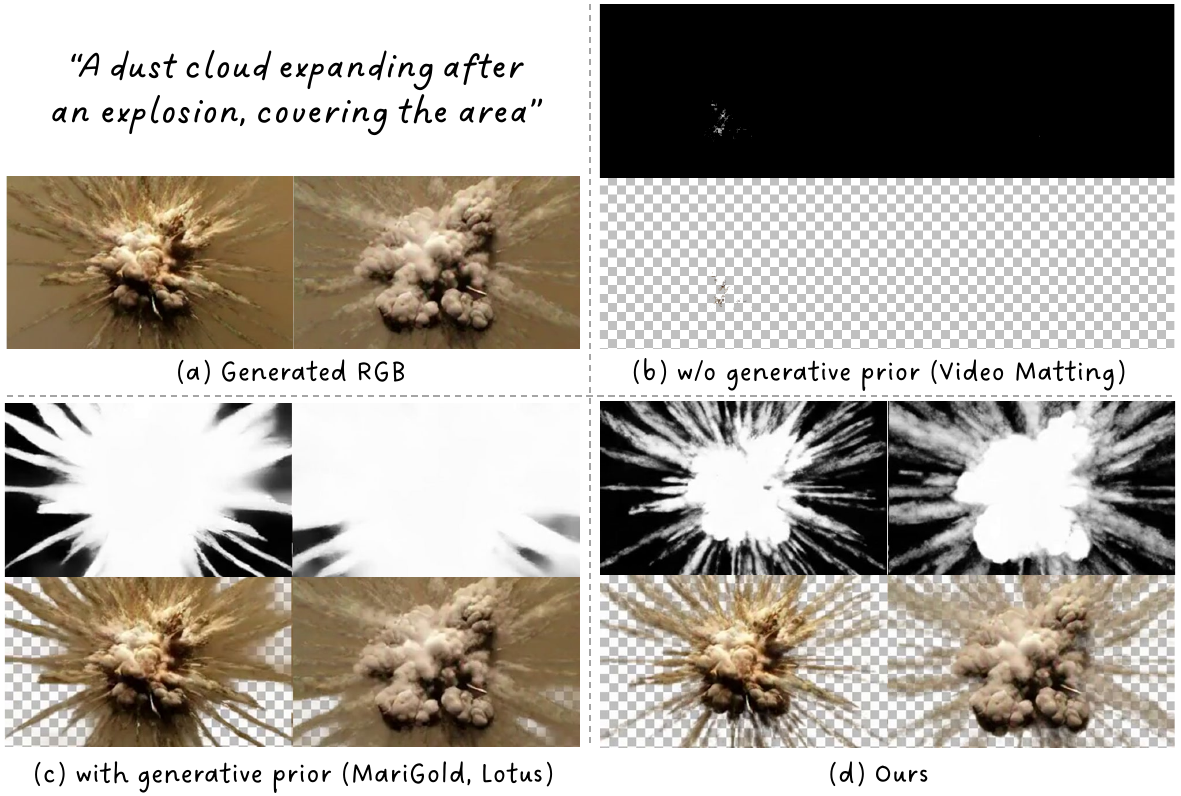}
    \vspace{-0.2in}
    \caption{Comparison between \textbf{Generation-Then-Prediction} and our \textbf{Joint Generation} approach. Given the generated RGB in (a), (b) and (c) show the predicted alpha (top) and the composited result (bottom). In (d), the top shows the jointly generated alpha.}
    \label{fig-intro}
    \vspace{-0.2in}
\end{figure}

Currently, no direct solutions exist for RGBA video generation, which remains a challenging task due to the scarcity of RGBA video data, with only around 484 videos available in~\cite{lin2021real}. This scarcity will significantly limit the diversity of generated content, resulting in a constrained set of object types and motion patterns.
One feasible solution is to use video matting~\cite{qin2023bimatting,lin2022robust,lin2023omnimatterf} to obtain alpha channels from generated videos. However, these methods are still limited by the scarcity of RGBA video data and struggle to generalize to a wider range of objects, as shown in Fig.~\ref{fig-intro} (b).
Other video segmentation methods, such as SAM-2~\cite{ravi2024sam2}, may generalize well to different tasks. However, they cannot generate alpha channels and are therefore unsuitable for direct compositing.
There have been attempts to generate RGBA at the image level, such as LayerDiffusion~\cite{zhang2024transparent}. 
However, adapting its concept directly to a temporal VAE used in video generative models remains challenging.

In this paper, we explore how to extend pretrained video models to generate corresponding alpha channels while retaining the original capabilities of pretrained models. Our goal is to generate content beyond the limitations of the current RGBA training set.
Existing works such as Lotus~\cite{he2024lotus} and Marigold~\cite{ke2024repurposing} demonstrate that leveraging pretrained generation model weights significantly enhances out-of-distribution in dense prediction, hinting at the potential for predicting alpha channels.
However, in the context of RGBA video generation, these approaches typically require generating RGB channels first, followed by separate alpha channel prediction. Consequently, information flows unidirectionally from RGB to alpha, keeping the two processes largely disconnected.
Given the limited availability of RGBA video data, this imbalance results in insufficient alpha prediction when challenging objects are generated, as shown in Fig.~\ref{fig-intro} (c).

In this work, we propose \textbf{TransPixeler}, which effectively adapts the pretrained RGB video models to generate RGB channels and the alpha channel simultaneously. 
We leverage state-of-the-art DiT-like video generation models~\cite{yang2024cogvideox, opensoraplan} 
, and additionally introduce new tokens appended after text and RGB tokens 
for generating the alpha channels. 
To facilitate convergence, we reinitialize the positional embeddings for the alpha tokens and introduce a zero-initialized, learnable domain embedding to distinguish alpha tokens from RGB tokens. Furthermore, we employ a LoRA-based fine-tuning scheme~\cite{hu2021lora}, applied exclusively to project alpha tokens into the qkv space, thereby maintaining RGB generation quality.
With the proposed approach, we extend the modality while preserving the original input-output structure and relying on the existing attention mechanism through LoRA adaptation. 

The extended sequence contains text, RGB, and alpha tokens, with self-attention divided into a 3x3 grouped attention matrix involving interactions like \textbf{Text-attend-to-RGB} (Text as query, RGB as key) and others. 
We also systematically analyze the attention mechanisms for RGBA generation: 1) \textbf{Text-attend-to-RGB} and \textbf{RGB-attend-to-Text}. The interaction between text and RGB tokens represents original model's generation capabilities. Minimizing the impact on text and RGB tokens during these attention computation processes can better retain the original model's performance;
2) \textbf{RGB-attend-to-Alpha}. We reveals a fundamental limitation in conventional methods is the lack of \textbf{RGB-attend-to-Alpha} attention. This attention is necessary to refine RGB tokens based on alpha information, improving RGB-alpha alignment; 3) \textbf{Text-attend-to-Alpha}. We remove this attention mechanism to reduce the risk caused by limited training data, which could degrade the model's performance. This removal also enhances the retention of the model's original capabilities.

By integrating these techniques, our method achieves diverse RGBA generation with limited training data while maintaining strong RGB-alpha alignment. To summarize, our contributions are as follows:
\begin{itemize}
    \item We propose an RGBA video generation framework using DiT models that requires limited data and training parameters, achieving diverse generation with strong alignment.
    \item We analyze the role of each attention component in the generation process, optimize their interactions, and introduce necessary modifications to improve RGBA generation quality.
    \item Our method is validated through extensive experiments, demonstrating its effectiveness across a variety of challenging scenarios.
\end{itemize}

\section{Related Work}
\label{sec:related}                             

\noindent{\textbf{Text-to-Video Generation.}}~Early video generation models were primarily based on Unet-based latent diffusion models (LDMs) extended from text-to-image models like Stable Diffusion~\cite{rombach2022high}. For example, AnimateDiff~\cite{guo2023animatediff} introduced a temporal attention module to improve temporal consistency across frames. Subsequent video generation models~\cite{wang2023modelscope, cerspense2023zeroscope, chen2023videocrafter1, chen2024videocrafter2, zhang2024moonshot, zhang2023show} adopted an alternating approach with 2D spatial and 1D temporal attention, including works like ModelScope, VideoCrafter, Moonshot, and Show-1. 

With advancements in large language models (LLMs) and the introduction of Sora~\cite{sora2024}, attention shifted from Unet architectures to transformer-based architectures (DiT). DiT-based video generation models, such as Latte~\cite{ma2024latte} and OpenSora~\cite{opensora}, extended the DiT text-to-image (T2I) model~\cite{chen2023pixart} and maintained the 2D and 1D alternating attention approach, achieving promising results. Recently, DiT-based video generation has rapidly progressed, achieving further improvements in quality. Several methods~\cite{yang2024cogvideox, opensoraplan, genmo2024mochi} have moved away from the 2D and 1D alternating approach, instead treating video frames as a single long sequence with 3D positional embeddings for encoding. These approaches also prepend text tokens—processed through a text encoder—to the video sequence, creating a streamlined network that relies solely on full self-attention and feed-forward layers. Our method builds upon these recent open-source transformer-based video generation models.

\vspace{0.5em}
\noindent{\textbf{Video Matting.}}~A straightforward approach for RGBA video generation is to extract the alpha channel from generated RGB content, as done with traditional green screen keying or learning-based video matting expert models~\cite{lin2023omnimatterf, lin2021real, lin2022robust}. OmnimatteRF~\cite{lin2023omnimatterf} introduces a video matting method that combines dynamic 2D foreground layers with a 3D background model, enabling more realistic scene reconstruction for real-world videos. Robust Video Matting (RVM)~\cite{lin2022robust} proposes a real-time, high-quality human video matting method with a recurrent architecture for improved temporal coherence, achieving state-of-the-art results without auxiliary inputs. Another work presents a high-speed, high-resolution background replacement technique with precise alpha matte extraction, supported by the VideoMatte240K and PhotoMatte13K/85 datasets~\cite{lin2021real}. Additionally, many image matting methods~\cite{chen2022pp, li2024matting, yao2024vitmatte, wang2024matting,burgert2024magick} can be applied for frame-by-frame matting.

Further, several works~\cite{he2024lotus, yang2024depth, ke2024repurposing} in image depth estimation adapt pretrained generation models for prediction tasks, achieving strong performance that often surpasses traditional, scratch-trained expert models. Marigold~\cite{ke2024repurposing} modifies architectures to create image-conditioned generation models, while Lotus~\cite{he2024lotus} explores the role of the diffusion process in this context. Although there is currently no dedicated approach for video matting within video generation models, we replicate and extend these methods to evaluate their performance, allowing us to highlight the limitations of prediction-based pipelines for RGBA generation.

\vspace{0.5em}
\noindent{\textbf{Generation beyond RGB.}}~Another category of methods~\cite{zhang2024transparent, long2024wonder3d, bao2023one, luo2024intrinsicdiffusion, zeng2024rgb, he2024lucidfusion, yang2023defect} explores expanding generation models to simultaneously generate additional channels, though they are not specifically designed for RGBA video generation. 
For instance, LayerDiffusion~\cite{zhang2024transparent} modifies the VAE in latent diffusion models to decode alpha channels. However, VAEs typically lack the semantic understanding required for precise alpha generation, limiting their effectiveness in complex visual scenarios where texture and contour details are critical. 
In contrast, other approaches~\cite{long2024wonder3d, bao2023one, luo2024intrinsicdiffusion, zeng2024rgb} modify the denoising model directly to enable joint generation. Wonder3D~\cite{long2024wonder3d} uses a domain embedding to control the model’s generation modality, while methods like IntrinsicDiffusion~\cite{luo2024intrinsicdiffusion} and RGB\(\leftrightarrow\)X~\cite{zeng2024rgb} adapt the UNet’s input and output layers to jointly produce intrinsic modalities. However, all these methods are designed for image tasks and rely on UNet architectures. When applied to video generation, they face limitations in quality and diversity due to the scarcity of RGBA video data.

\section{Method}
\label{sec:method}

\begin{figure*}[t]
    \centering
    \includegraphics[width=1.0\linewidth]{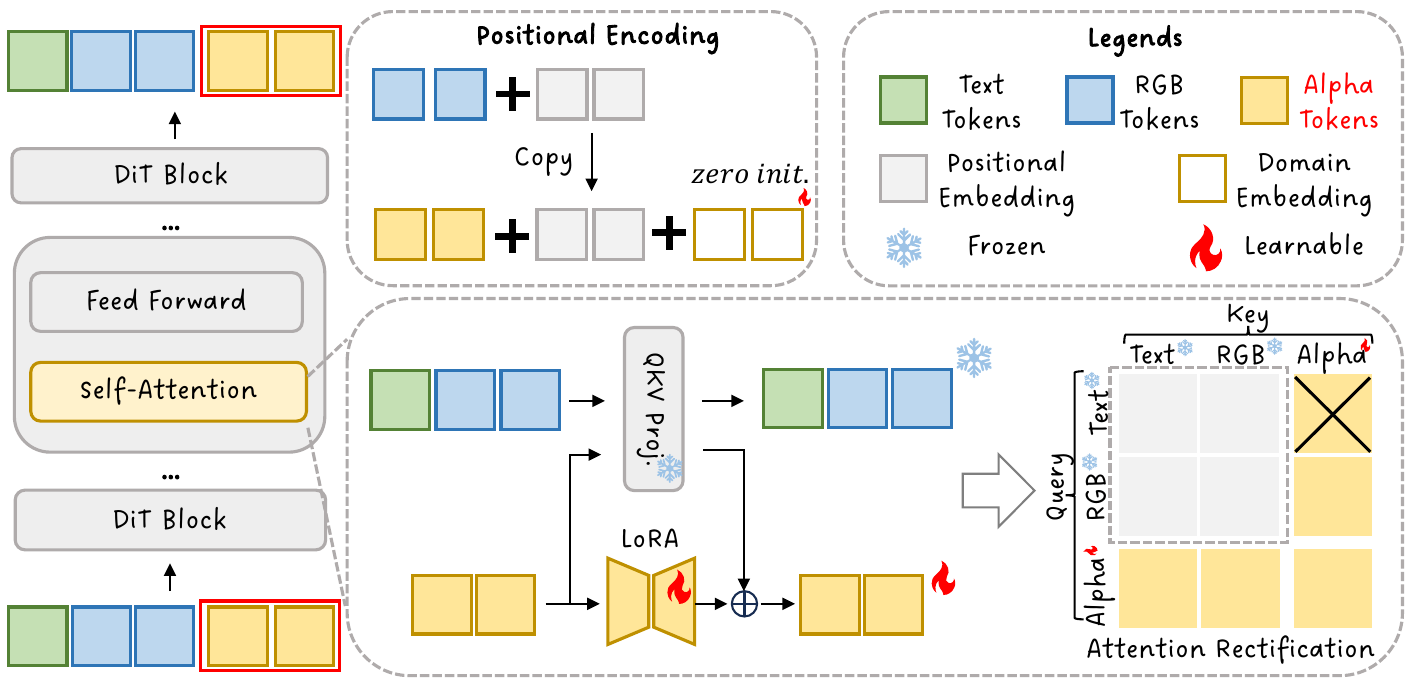}
    \vspace{-0.2in}
    \caption{\textbf{Pipeline of TransPixeler.} Our method is organized as follows: (1) \textbf{Left}: we extend the input of DiT to include new alpha tokens; (2) \textbf{Top Center}: we initialize alpha tokens with our positional encoding; (3) \textbf{Bottom Cente}r: we insert a partial LoRA and adjust attention computation during training and inference.}
    \label{fig-pipeline}
    \vspace{-0.1in}
\end{figure*}

\subsection{Preliminary}
We first introduce the open-sourced state-of-the-art DiT-based video generation models~\cite{yang2024cogvideox,genmo2024mochi}.
The core components of DiT-based video models are attention modules, and there are two primary distinctions between these models and previous approaches.
On one hand, unlike previous models that alternate between 1D temporal attention and 2D spatial attention~\cite{cerspense2023zeroscope, chen2023videocrafter1, chen2024videocrafter2, opensora}, current methods typically employ 3D spatio-temporal attention, allowing them to capture spatio-temporal dependencies more effectively.
On the other hand, instead of using cross-attention for text conditioning, these models concatenate text tokens \( \mathbf{x}_{\text{text}} \) with visual tokens \( \mathbf{x}_{\text{video}} \) into a single long sequence. 
The shape of video tokens and text tokens are \(B\times L\times D\) and \(B\times L_{\text{text}}\times D\), wher \(B\) equals to batch size, \(L_{\text{text}}\) equals to the length of text tokens, \(L\) equals to the length of video tokens and \(D\) equals to the latent dimension of transformer.
Full self-attention is then applied across the combined sequence:
\begin{equation}
\begin{aligned}
    &\text{Attention}(\mathbf{Q}, \mathbf{K}, \mathbf{V}) = \text{softmax}\left(\frac{\mathbf{Q}\mathbf{K}^T}{\sqrt{d_k}}\right)\mathbf{V}, \quad \text{where} \\
&\mathbf{Z : Z \in \{Q, K, V\}} \\&= [\mathbf{W}_{z : z \in \{q, k, v\}}(\mathbf{x}_{\text{text}}); \mathbf{f}_{z : z \in \{q, k, v\}}(\mathbf{x}_{\text{video}})]
\end{aligned}
\label{SA}
\end{equation}

Here \( \mathbf{W}_{t} \) (for \( t \in \{q, k, v\} \)) represents the projection matrixs in the transformer model, and \( \mathbf{f}_{t} \) (for \( t \in \{q, k, v\} \)) represents a combined operation that incorporates both the projection and positional encoding for visual tokens. 
There are two commonly used types of positional encoding. One is absolute positional encoding formulated as follows:
\begin{equation}
\begin{aligned}
\mathbf{f}_{z : z \in \{q, k, v\}}(\mathbf{x}_{\text{video}}) := \mathbf{W}_{z : z \in \{q, k, v\}}(\mathbf{x}_{\text{video}}^m + \mathbf{p}^m),
\end{aligned}
\label{PE}
\end{equation}
where \( \mathbf{p} \) is the positional embedding (e.g., a sinusoidal function) and \( m \) denotes the position of each RGB video token.
Another approach is the Rotary Position Embedding (RoPE)~\cite{su2024roformer}, often used by~\cite{yang2024cogvideox, genmo2024mochi}. 
This is expressed as
\begin{equation}
\begin{aligned}
\mathbf{f}_{z : z \in \{q, k\}}(\mathbf{x}_{\text{video}}) := \mathbf{W}_{z : z \in \{q, k\}}(\mathbf{x}_{\text{video}}^m) \circ e^{im\theta},
\end{aligned}
\label{RoPE}
\end{equation}
where \( m \) is the positional index, \( i \) is the imaginary unit for rotation, and \( \theta \) is the rotation angle.

\subsection{Our Approach} 
To jointly generate RGB and alpha videos, we adapt a pretrained RGB video generation model through several modifications. The whole pipeline is visualized in Fig.~\ref{fig-pipeline}.

Firstly, we double the sequence length of noisy input tokens to enable the model to generate videos of double length, from \( \mathbf{x}^{1:L}_{\text{video}} \) to \( \mathbf{x}^{1:2*L}_{\text{video}} \). 
Here, \( \mathbf{x}^{1:L}_{\text{video}} \) will be decoded into the RGB video, while \( \mathbf{x}^{L+1:2*L}_{\text{video}} \) will be decoded into the corresponding alpha video.
The Query(Q), Key(K), Value(V) representations are formulated as:
\begin{equation}
\begin{aligned}
&\mathbf{Z : Z \in \{Q, K, V\}} \\&= [\mathbf{W}_{z : z \in \{q, k, v\}}(\mathbf{x}_{\text{text}}); \mathbf{f}_{z : z \in \{q, k, v\}}(\mathbf{x}^{1:2*L}_{\text{video}})]
\end{aligned}
\end{equation}

In addition to sequence doubling, we explored increasing batch size or latent dimensions and splitting output into two domains; however, these approaches showed limited effectiveness under constrained datasets, which we discuss later.

\begin{figure}[t]
    \centering
    \includegraphics[width=1.0\linewidth]{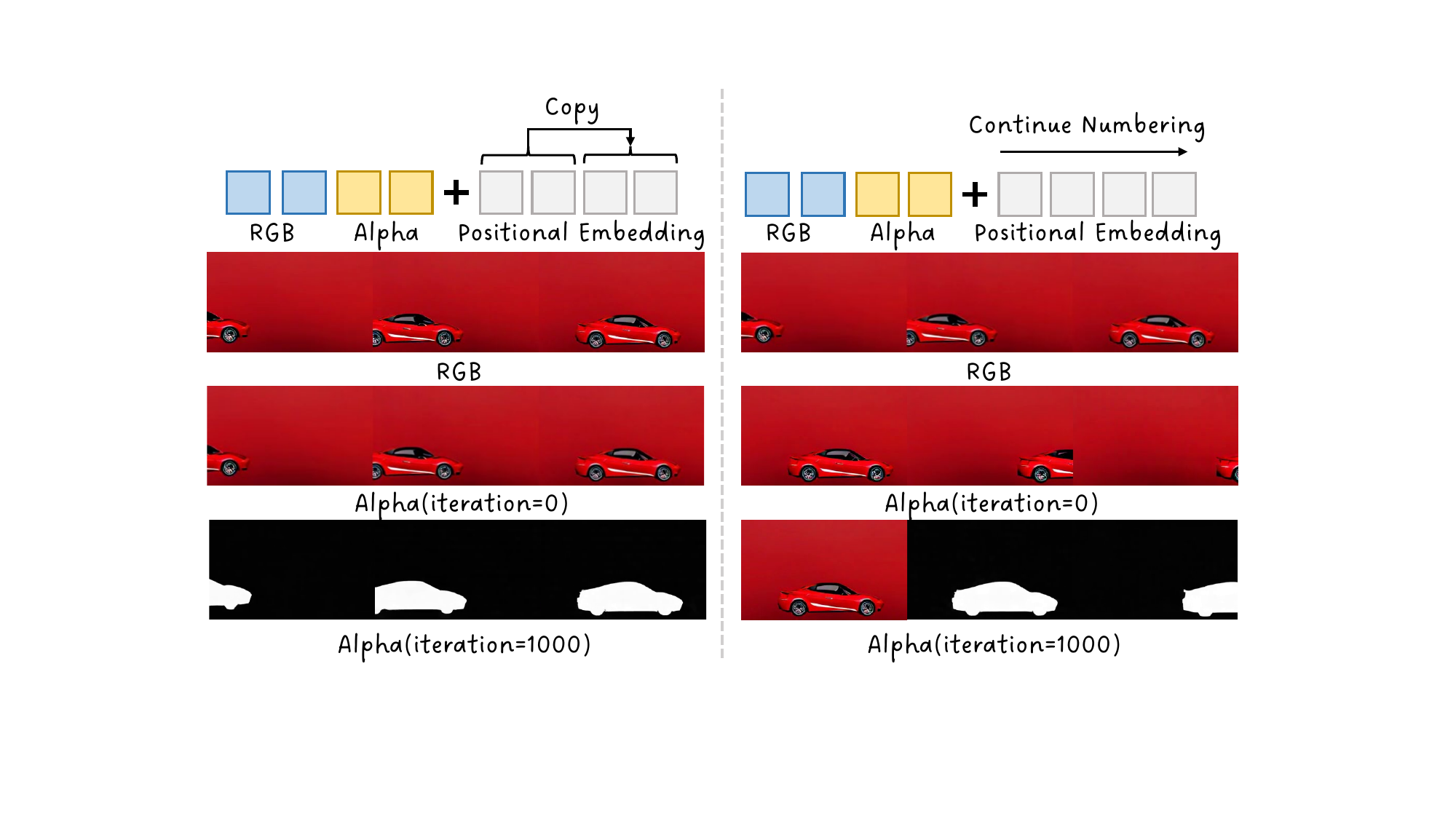}
    \vspace{-0.2in}
    \caption{\textbf{Positional Encoding Design for RGBA Generation.} Assigning alpha tokens the same positional encoding as RGB yields similar results, resulting in faster convergence after 1000 iterations compared to standard encoding strategies.}
    \label{fig-pe}
    \vspace{-0.1in}
\end{figure}

Secondly, we modify the positional encoding function \( \mathbf{f}_{t : t \in \{q, k, v\}}(\cdot) \), as shown in Fig.~\ref{fig-pe}.
Instead of continuously numbering indices, we allow RGB and alpha tokens to share the same positional encoding. 
Taking absolute positional encoding as an example:
\begin{equation}
\begin{aligned}
&\mathbf{f}^*_{z : z \in \{q, k, v\}}(\mathbf{x}_{\text{video}}) \\:=
&\begin{cases}
\mathbf{W}_{z : z \in \{q, k, v\}}(\mathbf{x}_{\text{video}}^m + \mathbf{p}^m), & \text{if } m \leq L, \\
\mathbf{W}^*_{z : z \in \{q, k, v\}}(\mathbf{x}_{\text{video}}^m + \mathbf{p}^{m-L} + d), & \text{if } m > L.
\end{cases}
\label{eq:our_pe}
\end{aligned}
\end{equation}

Here we introduce a domain embedding \( d \), initialized to zero. We make it learnable to help the model adaptively differentiate between RGB (\(m\leq L\)) and alpha tokens (\(m>L \)). 
The motivation behind this design is we observe that with same postional encoding, even initializing with different noises, the tokens from two domains tend to generate same results. 
It minimizes spatial-temporal alignment challenges at the very beginning of training and thus accelerates convergence.

Next we propose a fine-tuning scheme using LoRA~\cite{hu2021lora}, in which the LoRA layer is applied only to alpha domain tokens:
\begin{equation}
\begin{aligned}
&\mathbf{W}^*_{z : z \in \{q, k, v\}}(\mathbf{x}_{\text{video}}^m + \mathbf{p}^{m-L} + d)\\=
&\ \mathbf{W}_{z : z \in \{q, k, v\}}(\mathbf{x}_{\text{video}}^m + \mathbf{p}^{m-L} + d)
\\
+\ &\gamma\cdot \text{LoRA}(\mathbf{x}_{\text{video}}^m + \mathbf{p}^{m-L} + d), \quad \text{if } m > L,
\label{eq:our_lora}
\end{aligned}
\end{equation}
where \( \gamma \) controls the residual strength. 
Additionally, we design an attention mask to block unwanted attention computation. 
Given a text-video token sequence length \( L_\text{text} + 2L \), where \( L_\text{text} \) represents text token length, the mask is defined as:
\begin{equation}
\mathbf{M}^*_{mn} = 
\begin{cases} 
-\infty, & \text{if } m \leq L_\text{text} \ \text{and} \ \, n > L_\text{text} + L, \\
0, & \text{otherwise}.
\end{cases}
\label{eq:attn_mask}
\end{equation}

Combining these modifications, inference with our method is expressed as:
\begin{equation}
\begin{aligned}
    &\text{Attention}(\mathbf{Q}, \mathbf{K}, \mathbf{V}) = \text{softmax}\left(\frac{\mathbf{Q}\mathbf{K}^T}{\sqrt{d_k}}+\mathbf{M}^*\right)\mathbf{V}, \quad \text{where} \\
&\mathbf{Z : Z \in \{Q, K, V\}} \\&= [\mathbf{W}_{z : z \in \{q, k, v\}}(\mathbf{x}_{\text{text}}); \mathbf{f}^*_{z : z \in \{q, k, v\}}(\mathbf{x}_{\text{video}})]
\end{aligned}
\label{eq:our_method}
\end{equation}

Training is carried out using flow matching~\cite{liu2022flow} or a traditional diffusion process~\cite{ho2020denoising}.

\subsection{Analysis}
Given our goal of maximizing the inherited capabilities of the pretrained video model, enabling it to generate beyond the existing RGBA training set, we analyze the most critical component within our current 3D full attention DiT video generation model: the attention mechanism.
The attention matrix, \(\mathbf{Q}\mathbf{K}^T\), has dimensions \((L_\text{text} + 2*L) \times (L_\text{text} + 2*L)\), which we simplify by organizing it into a 3x3 grouped attention matrix—including \textbf{Text-attend-to-RGB}, \textbf{RGB-attend-to-Text}, and so forth, as illustrated in Fig.~\ref{fig-pipeline}. 

\vspace{0.5em}
\noindent\textbf{Text-Attend-to-RGB and RGB-Attend-to-Text}. These represent the upper-left 2x2 section of  and are computations that exist solely in the original RGB generation model. If we ensure that this part of the computation remains unaffected, we can replicate the original RGB generation performance. Therefore, we limit the scope of LoRA's influence, as defined in Eq.~\eqref{eq:attn_mask}, by retaining the original QKV values for both text and RGB tokens, thus preserving the pretrained model’s behavior in these domains.

\begin{figure}[t]
    \centering
    \includegraphics[width=1.0\linewidth]{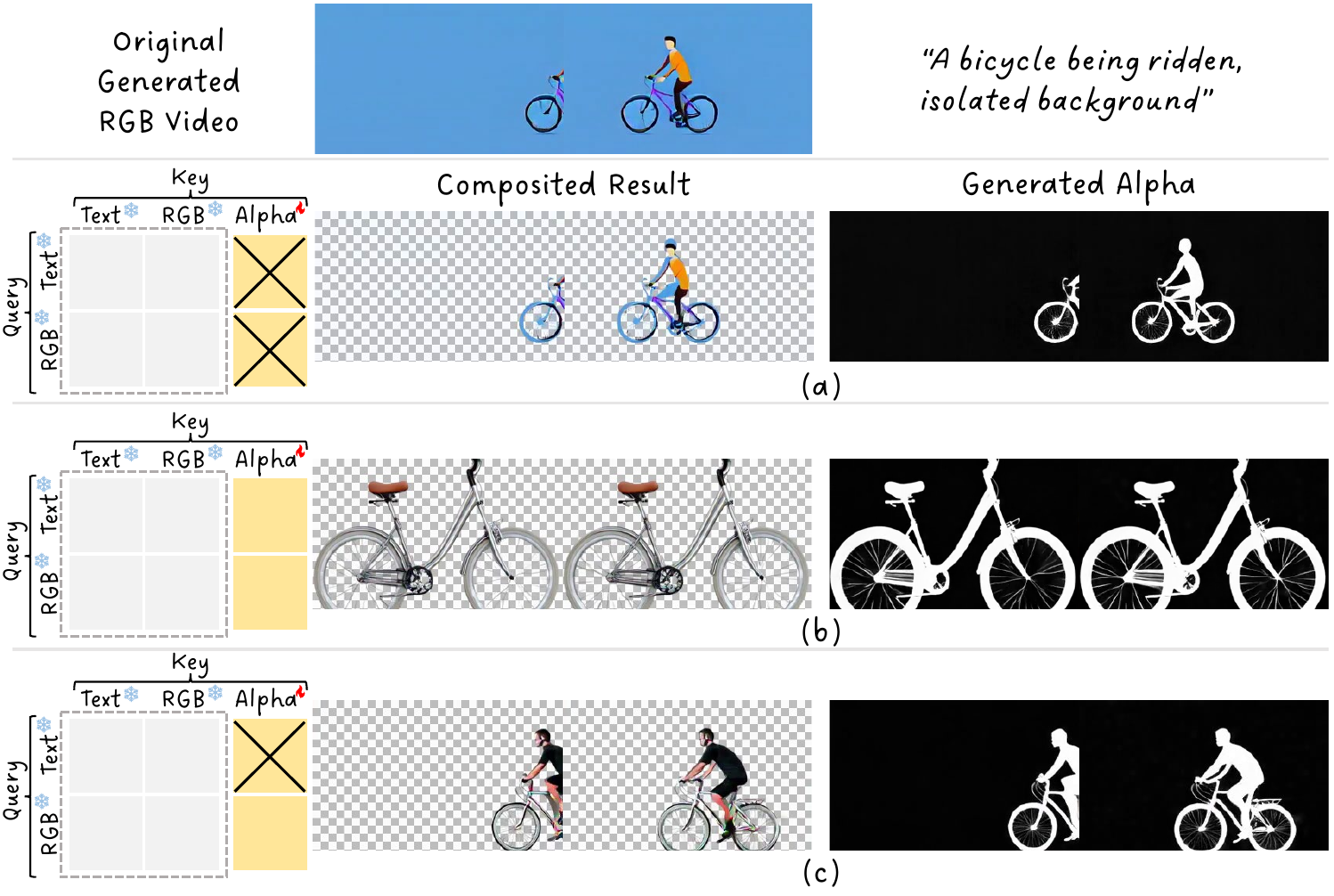}
    \vspace{-0.2in}
    \caption{\textbf{Attention Rectification.} (a) Eliminating all attention from alpha as a key preserves 100\% RGB generation but leads to poor alignment. (b) Retaining all attention significantly degrades quality, causing a lack of motion in bicycles. (c) Our method achieves an effective balance.
    }
    \label{fig-attn}
    \vspace{-0.1in}
\end{figure}

Besides the partial LoRA, the added alpha tokens requires the text and RGB tokens to also act as queries and interact with the alpha tokens as keys, which alters the computation in this 2x2 attention matrix. 
Therefore, we further analyze two additional attention computations that impact RGB generation, as shown in Fig.~\ref{fig-attn}.

\vspace{0.5em}
\noindent\textbf{Text-Attend-to-Alpha.} We find that this attention is detrimental to the generation quality. Since the model was originally trained with text and RGB data, introducing attention from text to alpha causes interference due to the domain gap between alpha and RGB. Specifically, the alpha modality provides only contour information and lacks the rich texture, color, and semantic details associated with the text prompt, thereby degrading generation quality. To mitigate this, we design the attention mask (Eq.~\eqref{eq:attn_mask}) that blocks this computation.

\vspace{0.5em}
\noindent\textbf{RGB-Attend-to-Alpha.} In contrast, we identify \textbf{RGB-to-Alpha} as essential for successful joint generation. This attention allows the model to refine RGB tokens by considering alpha information, facilitating alignment between generated RGB and alpha channels. This refinement process is a critical component missing in previous generation-then-prediction pipelines, which lacked a feedback mechanism for RGB refinement based on alpha guidance.

\begin{figure*}[htbp]
    \centering
    \includegraphics[width=1.0\linewidth]{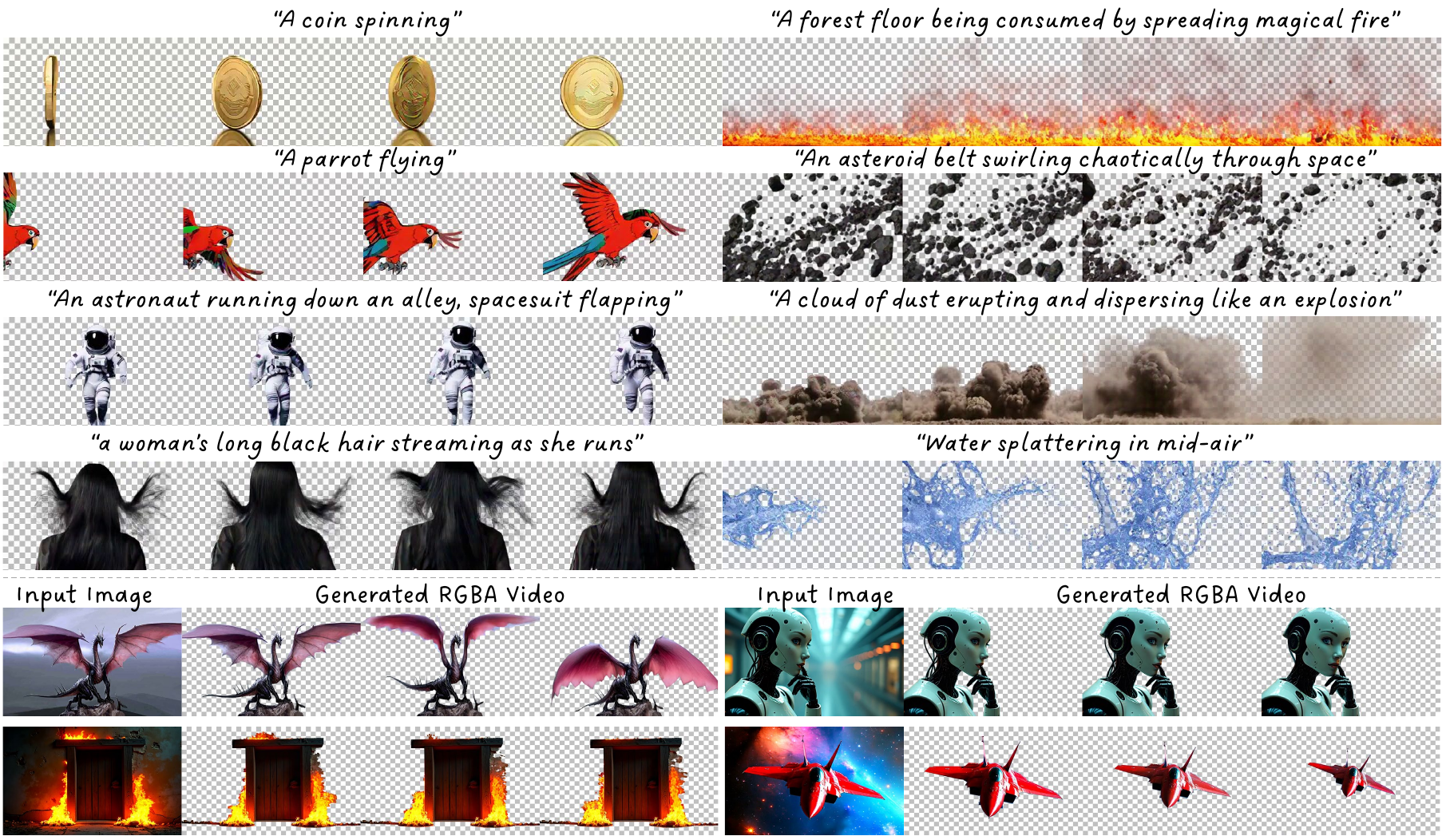}
    \vspace{-0.2in}
    \caption{\textbf{Applications.} \textbf{Top}: Text-to-Video with Transparency. \textbf{Bottom}: Image-to-Video generation with transparency.
.}
    \label{fig-applications}
    \vspace{-0.1in}
\end{figure*}

\section{Experiment}
\label{sec:exp}


\noindent\textbf{Training Dataset.}~We utilize the public VideoMatte240K dataset~\cite{lin2021real}, a comprehensive collection of 484 high-resolution green screen videos consists of 240,709 unique frames of alpha mattes and foregrounds. These frames provide a diverse range of human subjects, clothing styles, and poses. We apply fundamental preprocessing steps for them, including color decontamination and background blurring. Prompts are extracted using ShareGPT4V~\cite{chen2023sharegpt4v}. 

\vspace{0.5em}
\noindent\textbf{Model.}~Our RGBA video diffusion models are developed by fine-tuning pre-trained diffusion models. Specifically, we employ two models based on the diffusion transformer architecture: the open-source model CogVideoX~\cite{yang2024cogvideox} and a modified variant of CogVideoX denoted as \(J\). 
CogVideoX generates RGB videos at a resolution of 480x720 with 49 frames at 8 FPS, using 50 sampling steps. In contrast, the modified version produces videos at a resolution of 176x320 with 64 frames at 24 FPS, while also using 50 sampling steps. Additionally, we integrate our method with CogVideoX-I2V (Image-to-Video) to support image-to-video generation with transparency.
We set the LoRA rank to 128. For domain embedding, we initialize it with an original shape of \(1\times D\) and zero values, then expand it to \(L\times D\) through repetition during training. 
We train these parameters over 5,000 iterations with a batch size of 8 in total, utilizing 8 NVIDIA A100 GPUs.

\begin{figure*}[htbp]
    \centering
    \includegraphics[width=1.0\linewidth]{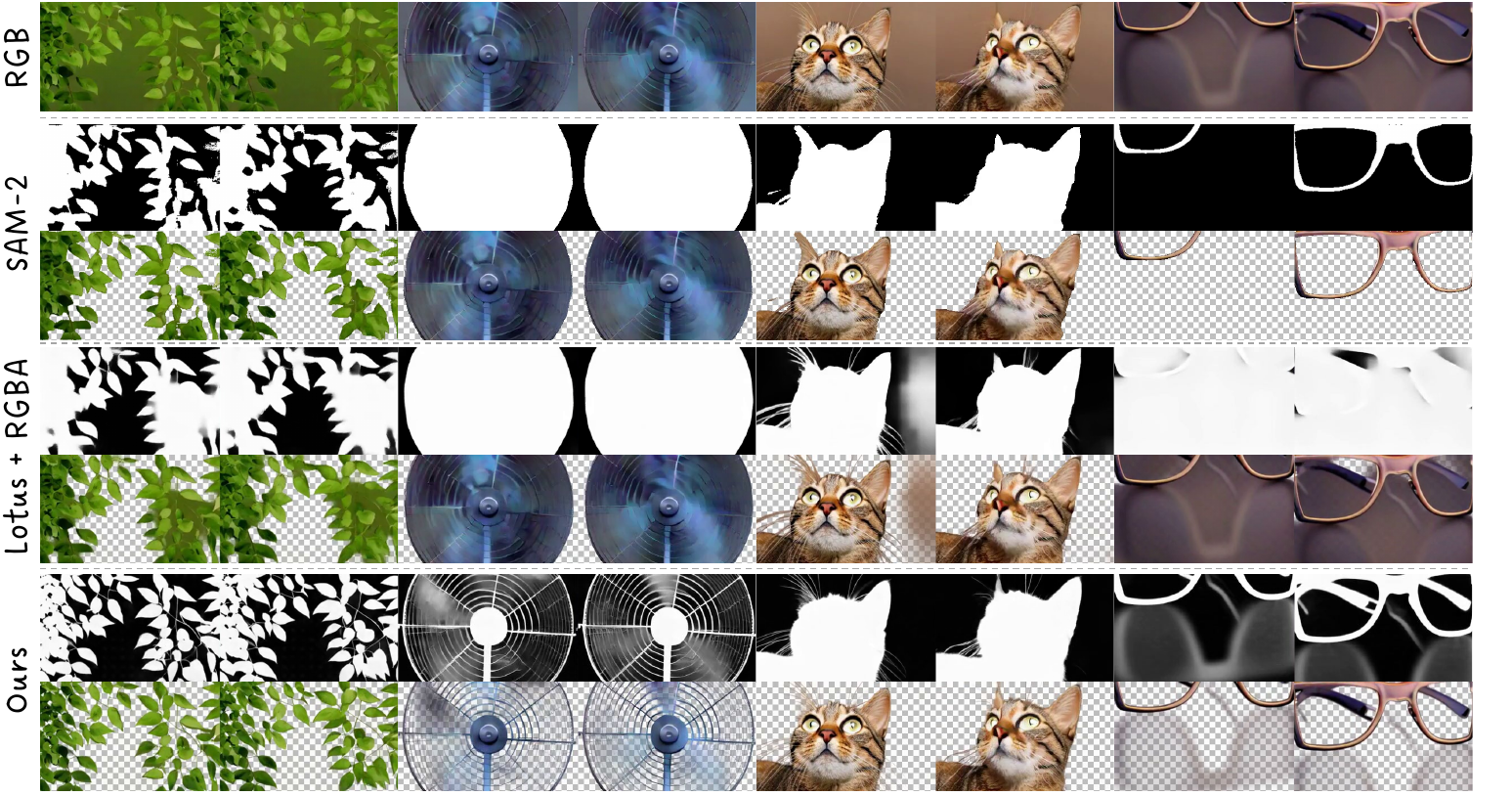}
    \vspace{-0.2in}
    \caption{\textbf{Comparison with Generation-then-Prediction Pipelines.} Our method demonstrates superior alignment.
}
    \label{fig-comparison}
\end{figure*}

\begin{figure*}[htbp]
    \centering
    \includegraphics[width=1.0\linewidth]{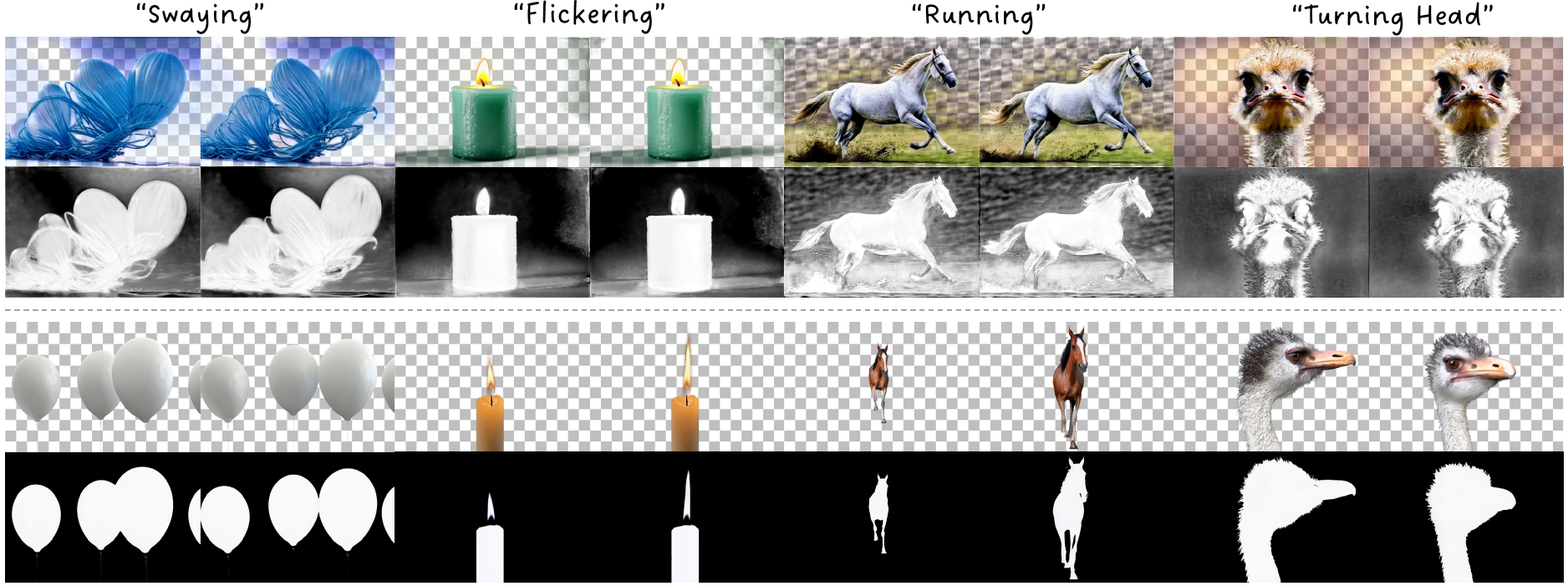}
    \caption{\textbf{Comparison with Joint Generation Pipelines.} \textbf{Top}: LayerDiffusion + AnimateDiff; \textbf{Bottom}: Ours. Our method achieves better alignment and generates corresponding motion described by prompts.}
    \label{fig-comparison-generation}
    \vspace{-0.1in}
\end{figure*}

\subsection{Applications}
We mainly demonstrate two applications shown in Fig.~\ref{fig-applications}: 

\vspace{0.5em}
\noindent\textbf{Text-to-Video with Transparency.}~Our method is capable of generating moving objects with various types of motion, such as spinning, running, and flying, while also handling transparent properties of bottles and glasses. Additionally, it can produce complex visual effects, including fire, explosions, cracking, and lightning, as well as creative examples.

\vspace{0.5em}
\noindent\textbf{Image-to-Video with Transparency.}~Our method can also be integrated with an I2V video generation model-CogVideoX-I2V. Users can provide a single image along with an alpha channel (optional), and then we generate subsequent frames with dynamic effects and automatically propagate or generate alpha channels for these frames.

\subsection{Comparisons}

\noindent\textbf{Generation-then-Prediction Pipeline.}~As shown in Fig.~\ref{fig-intro}, video matting methods~\cite{lin2021real,qin2023bimatting,yao2024matte} struggle with matting non-human objects (see supplementary materials for additional results). Therefore, we selected Lotus~\cite{he2024lotus} and SAM-2~\cite{ravi2024sam2} as baselines due to their stronger generalization: Lotus uses pretrained generative models, and SAM-2 is trained on large datasets. Since Lotus was originally designed for single-image depth estimation, we extended it for RGBA videos, denoted as Lotus + RGBA in our comparisons. Qualitative results are shown in Fig.~\ref{fig-comparison}. Since ground-truth alpha channels are not available for generated videos, we focus on qualitative comparison.

\vspace{0.5em}
\noindent\textbf{Joint Generation Pipeline.}~Since there are currently no existing RGBA video generation models, we integrate AnimateDiff~\cite{guo2023animatediff} with LayerDiffusion~\cite{zhang2024transparent} to generate RGBA videos. We use the open-source video generation model CogVideoX~\cite{yang2024cogvideox} as the base model for fair comparison. 
The qualitative results are illustrated in Fig.~\ref{fig-comparison-generation}.

\vspace{0.5em}
\noindent\textbf{User Study.}~
We also conduct a user study with Amazon Mechanical Turk to compare two joint generation methods, as shown in Table.~\ref{tab:user_study}. Participants are asked to evaluate two key aspects: 1) whether the RGB and alpha align correctly; and 2) whether the motion in the generated video matches the corresponding text description. A total of 30 videos are generated from distinct text prompts, and 87 users participated in the evaluation. The study shows that our method is obviously favored more by users with higher votes.

\begin{table}
    \centering
    \caption{\textbf{User Study.}}
    \vspace{-0.1in}
    \resizebox{1.0 \columnwidth}{!}{%
    \begin{tabular}{ccc} \toprule
         &RGBA Alignment  &Motion Quality \\ \toprule AnimateDiff~\cite{guo2023animatediff}+LayerDiff~\cite{zhang2024transparent}& 6.7\% &21.7\% \\ 
         Ours + CogVideoX~\cite{yang2024cogvideox}& \textbf{93.3\%} &\textbf{78.3\%} \\ \bottomrule
    \end{tabular}
    }
    \vspace{-0.1in}
    \label{tab:user_study}
\end{table}

\subsection{Ablation Study}
\begin{figure}[htbp]
    \centering
    \includegraphics[width=1.0\linewidth]{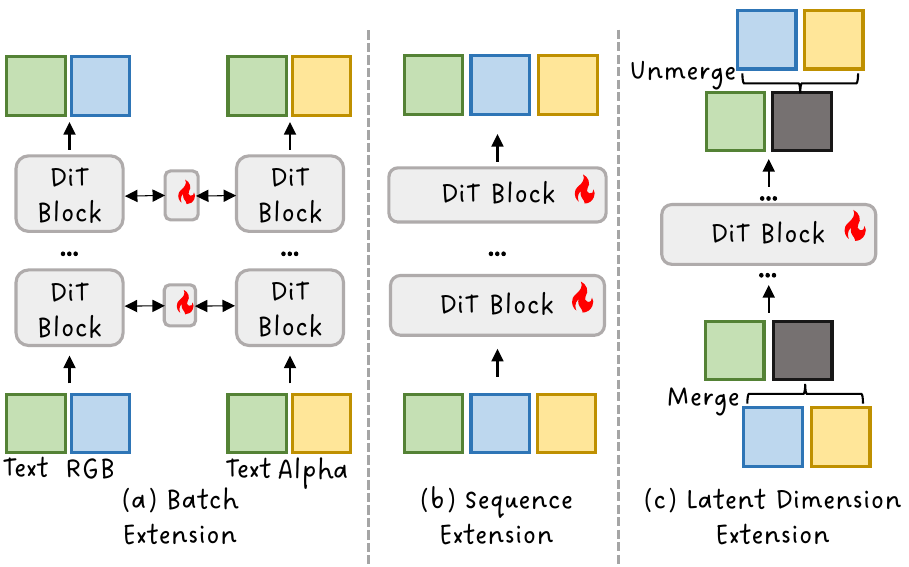}
    \caption{\textbf{Alternative Designs for Joint Generation with DiT}. Sequence extension (b) represents our method.}
    \label{fig-arch}
\end{figure}
\begin{figure}[h]
    \centering
    \includegraphics[width=1.0\linewidth]{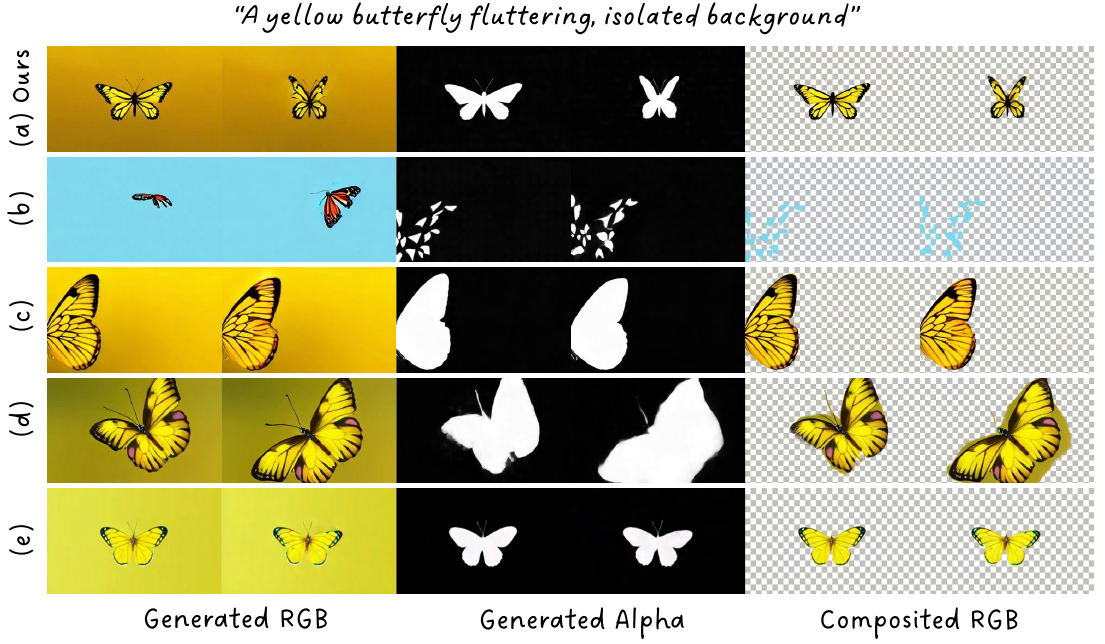}
    \caption{\textbf{Ablation Study.} (a) Ours; (b) Ours without RGB-attend-to-Alpha; (c) Ours with Text-attend-to-alpha; (d) Batch Extension Strategy; (e) Latent Dimension Extension Strategy. Our method maintains high-quality motion generation (e.g., butterflies waving their wings) while achieving good alignment.}
    \label{fig-ablation}
\end{figure}

As shown in Fig.~\ref{fig-ablation}, we conduct the ablation study across two dimensions: attention rectification and network design.

\vspace{0.5em}
\noindent\textbf{Attention Rectification.}~
By blocking RGB-to-Alpha attention, we first validate the importance of RGB-to-Alpha attention for aligning RGB and alpha channels, a feature lacking in most prediction-based methods. We also examine the effect of removing unnecessary attention to preserve the model's generative capacity, by learning Text-to-Alpha attention only.
Without RGB-to-Alpha attention, the alpha channel misaligns with RGB content 
and the RGB output loses motion quality (e.g., reverse rocket).

\vspace{0.5em}
\noindent\textbf{Alternative Designs For Joint Generation.}~
Given the transformer’s input dimensions \( B \times L \times D \), we extend the sequence dimension \( L \) to produce RGB and alpha channels, but alternative extensions are possible at the Batch \( B \) or Latent Dimension \( D \) levels (see Fig.~\ref{fig-arch}). In the \textbf{Batch Extension} approach, a new module enables inter-batch communication, similar to the technique in~\cite{vainer2024collaborative}. For \textbf{Latent Dimension Extension}, we merge video and alpha tokens, project them into the DiT model’s latent space, and unmerge post-generation, using learnable linear layers with fine-tuning. Batch Extension shows weaker RGB-alpha alignment, while Latent Dimension Extension, though akin to training from scratch, significantly reduces diversity.

\begin{figure}[t]
    \centering
    \includegraphics[width=0.9\linewidth]{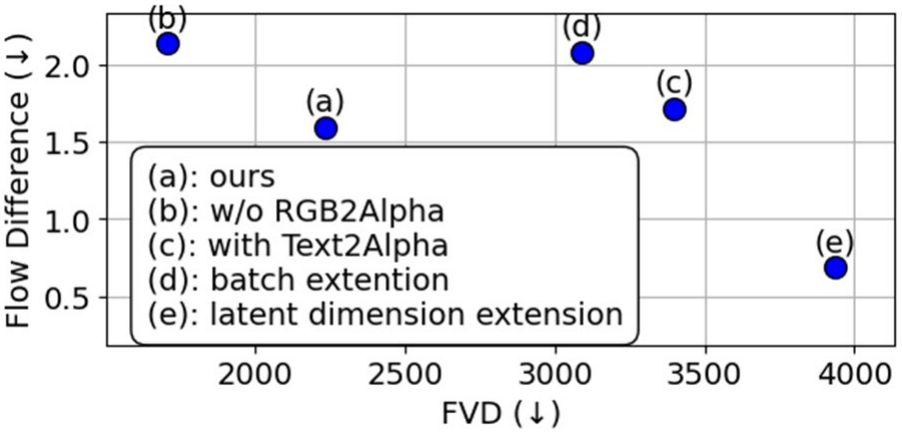}
    \vspace{-0.1in}
    \caption{\textbf{Quantitative Evaluation}. Our approach achieves a good balance between alignment (low flow difference) and preserving generative quality (low FVD).}
    \label{fig-abaltion-metrics}
    \vspace{-0.1in}
\end{figure}

\vspace{0.5em}
\noindent\textbf{Evaluation.}~
In addition to the qualitative comparisons shown in Fig.~\ref{fig-ablation}, we also generated a total of 80 videos, each consisting of 64 frames, and evaluated them using two primary metrics:
\textbf{Flow Difference.} To measure alignment between the generated RGB and Alpha videos, we use optical flow~\cite{horn1981determining} to focus on motion consistency while ignoring appearance. Specifically, we calculate optical flow with Farneback method~\cite{farneback2003two} and compute the flow difference as the average Euclidean distance between RGB and Alpha flow fields.
\textbf{Frechét Video Distance (FVD).} We use FVD~\cite{unterthiner2019fvd} to compare the RGB videos generated by each RGBA method against those from the original RGB model, evaluating how well each method preserves the model's original generative quality. A lower FVD indicates that the generated results are closer to the original RGB model in terms of motion coherence and diversity, thus demonstrating a high fidelity to the model's intended generative quality.
Results are shown in Fig.~\ref{fig-abaltion-metrics}.

\section{Conclusion}
\label{sec:conclusion}
In this work, we present a novel approach for Text-to-RGBA video generation, extending RGB generation models to support RGBA output with minimal modification and high fidelity. By leveraging transformer-based DiT models and optimizing attention mechanisms specific to RGBA generation, our method effectively balances the preservation of RGB quality with the accurate generation of alpha channels. Our approach demonstrates that targeted modifications—such as the addition of alpha tokens, reinitialization of positional embeddings, and selective LoRA fine-tuning—can yield complex and high-quality RGBA outputs even with limited data. Extensive experimental results validate our framework, showing its versatility and robustness across diverse scenarios. Looking forward, we aim to explore further optimizations to reduce computational costs and enhance model scalability.
\clearpage
\setcounter{page}{1}
\maketitlesupplementary
\section{Limitations}
\label{sec:limitations}
Our DiT-based method for RGBA generation incurs quadratic computational costs due to sequence expansion. However, our method achieves an optimal balance between generation and alignment when trained with a limited dataset. Numerous studies~\cite{wang2020linformer, ding2023longnet, zhu2021long} have addressed the computational overhead of long sequences, with many optimizations reducing complexity to a linear scale. To enhance the efficiency of our method, we plan to incorporate these optimizations in future work. Additionally, our performance is influenced by the generative priors provided by the chosen T2V model, which affects the quality and consistency of our outputs.

\section{Comparisons with Video Matting}
\label{sec:comparisons}
We compare our method with video matting methods BiMatting~\cite{qin2023bimatting} and Robust Video Matting (RVM)~\cite{lin2022robust}, as well as the image matting method Matte-Anything~\cite{yao2024matte}. From the results, it is evident that most methods, trained on the VideoMatte240k~\cite{lin2021real} dataset, struggle to produce valid outputs for non-human objects, often resulting in empty results. Even image matting methods trained on large-scale datasets fail to handle certain visual effects correctly. Results are shown in the attached HTML source files.

\section{Data Preprocessing}
\label{sec:data-preprocessing}
\noindent\textbf{Color Decontamination}. In our method, we preprocess the training data by applying a color decontamination step to enhance the quality of the RGBA video generation. Color contamination typically occurs when there is an undesired blending of foreground and background colors, especially along the edges of an object, due to imperfect alpha masks. This blending causes color bleeding, where the foreground and background colors mix, resulting in lower quality RGBA frames with inaccurate color representation. To address this issue, we refine the alpha mask using parameters such as gain ($\gamma=1.1$) and choke ($\chi=0.5$) to adjust the sharpness and influence of the mask edges. The decontaminated RGB values are then computed as follows:

\[
\text{RGB}_{\text{decon}} = \text{RGB} \times (1 - \text{mask}_{\text{refined}}) + \text{mask}_{\text{refined}} \times \text{Background}
\]

This equation ensures that unwanted color contamination is minimized, providing a more precise distinction between foreground and background regions. By performing this preprocessing step, we generate high-quality training data that significantly improves the performance of our RGBA video generation model.

\noindent\textbf{Background Blurring}. Unlike typical training strategies in video matting methods, where objects are composited with complex backgrounds to increase the difficulty of the task, our goal is to support joint generation of alpha and RGB channels while ensuring alignment between them. Instead of emphasizing complex matting, we focus on generating consistent and high-quality output by compositing objects with simple, static backgrounds that match the black areas in the alpha channel. Specifically, we apply a large Gaussian blur kernel of size 201 to the first frame to create a blurred background and blend each subsequent frame with this static background. This approach helps simplify the training conditions, allowing the model to better align the RGB and alpha components while maintaining high-quality output.

\section{Optical Flow Difference}
\label{sec:optical-flow}
To evaluate the alignment between the RGB and alpha channels in generated videos, we introduce a metric based on optical flow difference. Optical flow measures the apparent motion of objects between consecutive frames, and comparing the optical flow fields of RGB and alpha frames provides insight into the consistency of motion across these modalities. Specifically, we use the Farneback method (\texttt{cv::calcOpticalFlowFarneback}) to compute the optical flow for both RGB and alpha frames, and then calculate the average Euclidean distance between their flow vectors as a measure of misalignment. This approach quantifies the degree to which the RGB and alpha channels align in terms of motion.

\noindent\textbf{Pseudo Code Overview}:
\begin{enumerate}
    \item \textbf{Load consecutive RGB and alpha frames} from the input video.
    \item \textbf{Convert the frames to grayscale} for optical flow computation, as optical flow is typically calculated on intensity values.
    \item \textbf{Compute optical flow using the Farneback method} (\texttt{cv::calcOpticalFlowFarneback}) for both the RGB and alpha frames.
    \item \textbf{Calculate the Euclidean distance} between the RGB and alpha flow vectors for each pixel.
    \item \textbf{Average the differences} across all pixels and frames to obtain the final optical flow difference.
\end{enumerate}

The average optical flow difference provides a quantitative metric for evaluating the alignment between RGB and alpha channels, helping to ensure that both modalities exhibit consistent motion.

\section{Video Results}
\label{sec:video-results}
For all video results shown in the main paper, please see the attached HTML source files.

\section{Additional Visual Results}
\label{sec:additional-visual-results}
In addition to the video results in the main paper, we provide more generated results in the supplementary files, including various objects and visual effects. Please find the corresponding results in the supplementary files.

\newpage
{
    \small
    \bibliographystyle{ieeenat_fullname}
    \bibliography{main}

\begin{thebibliography}{67}
\providecommand{\natexlab}[1]{#1}
\providecommand{\url}[1]{\texttt{#1}}
\expandafter\ifx\csname urlstyle\endcsname\relax
  \providecommand{\doi}[1]{doi: #1}\else
  \providecommand{\doi}{doi: \begingroup \urlstyle{rm}\Url}\fi

\bibitem[Bao et~al.(2023)Bao, Nie, Xue, Li, Pu, Wang, Yue, Cao, Su, and Zhu]{bao2023one}
Fan Bao, Shen Nie, Kaiwen Xue, Chongxuan Li, Shi Pu, Yaole Wang, Gang Yue, Yue Cao, Hang Su, and Jun Zhu.
\newblock One transformer fits all distributions in multi-modal diffusion at scale.
\newblock In \emph{International Conference on Machine Learning}, pages 1692--1717. PMLR, 2023.

\bibitem[Blattmann et~al.(2023)Blattmann, Dockhorn, Kulal, Mendelevitch, Kilian, Lorenz, Levi, English, Voleti, Letts, et~al.]{blattmann2023stable}
Andreas Blattmann, Tim Dockhorn, Sumith Kulal, Daniel Mendelevitch, Maciej Kilian, Dominik Lorenz, Yam Levi, Zion English, Vikram Voleti, Adam Letts, et~al.
\newblock Stable video diffusion: Scaling latent video diffusion models to large datasets.
\newblock \emph{arXiv preprint arXiv:2311.15127}, 2023.

\bibitem[Brooks et~al.(2024)Brooks, Peebles, Holmes, DePue, Guo, Jing, Schnurr, Taylor, Luhman, Luhman, Ng, Wang, and Ramesh]{sora2024}
Tim Brooks, Bill Peebles, Connor Holmes, Will DePue, Yufei Guo, Li Jing, David Schnurr, Joe Taylor, Troy Luhman, Eric Luhman, Clarence Ng, Ricky Wang, and Aditya Ramesh.
\newblock Video generation models as world simulators.
\newblock 2024.

\bibitem[Burgert et~al.(2024)Burgert, Price, Kuen, Li, and Ryoo]{burgert2024magick}
Ryan~D Burgert, Brian~L Price, Jason Kuen, Yijun Li, and Michael~S Ryoo.
\newblock Magick: A large-scale captioned dataset from matting generated images using chroma keying.
\newblock In \emph{Proceedings of the IEEE/CVF Conference on Computer Vision and Pattern Recognition}, pages 22595--22604, 2024.

\bibitem[cerspense(2023)]{cerspense2023zeroscope}
cerspense.
\newblock zeroscope\_v2.
\newblock \url{https://huggingface.co/cerspense/zeroscope_v2_576w}, 2023.
\newblock Accessed: 2023-02-03.

\bibitem[Chen et~al.(2022)Chen, Liu, Wang, Peng, Hao, Chu, Tang, Wu, Chen, Yu, et~al.]{chen2022pp}
Guowei Chen, Yi Liu, Jian Wang, Juncai Peng, Yuying Hao, Lutao Chu, Shiyu Tang, Zewu Wu, Zeyu Chen, Zhiliang Yu, et~al.
\newblock Pp-matting: high-accuracy natural image matting.
\newblock \emph{arXiv preprint arXiv:2204.09433}, 2022.

\bibitem[Chen et~al.(2023{\natexlab{a}})Chen, Xia, He, Zhang, Cun, Yang, Xing, Liu, Chen, Wang, et~al.]{chen2023videocrafter1}
Haoxin Chen, Menghan Xia, Yingqing He, Yong Zhang, Xiaodong Cun, Shaoshu Yang, Jinbo Xing, Yaofang Liu, Qifeng Chen, Xintao Wang, et~al.
\newblock Videocrafter1: Open diffusion models for high-quality video generation.
\newblock \emph{arXiv preprint arXiv:2310.19512}, 2023{\natexlab{a}}.

\bibitem[Chen et~al.(2024)Chen, Zhang, Cun, Xia, Wang, Weng, and Shan]{chen2024videocrafter2}
Haoxin Chen, Yong Zhang, Xiaodong Cun, Menghan Xia, Xintao Wang, Chao Weng, and Ying Shan.
\newblock Videocrafter2: Overcoming data limitations for high-quality video diffusion models.
\newblock \emph{arXiv preprint arXiv:2401.09047}, 2024.

\bibitem[Chen et~al.(2023{\natexlab{b}})Chen, Yu, Ge, Yao, Xie, Wu, Wang, Kwok, Luo, Lu, et~al.]{chen2023pixart}
Junsong Chen, Jincheng Yu, Chongjian Ge, Lewei Yao, Enze Xie, Yue Wu, Zhongdao Wang, James Kwok, Ping Luo, Huchuan Lu, et~al.
\newblock Pixart-alpha: Fast training of diffusion transformer for photorealistic text-to-image synthesis.
\newblock \emph{arXiv preprint arXiv:2310.00426}, 2023{\natexlab{b}}.

\bibitem[Chen et~al.(2023{\natexlab{c}})Chen, Li, Dong, Zhang, He, Wang, Zhao, and Lin]{chen2023sharegpt4v}
Lin Chen, Jisong Li, Xiaoyi Dong, Pan Zhang, Conghui He, Jiaqi Wang, Feng Zhao, and Dahua Lin.
\newblock Sharegpt4v: Improving large multi-modal models with better captions.
\newblock \emph{arXiv preprint arXiv:2311.12793}, 2023{\natexlab{c}}.

\bibitem[Chen et~al.(2023{\natexlab{d}})Chen, Ji, Wu, Wu, Xie, Li, Xia, Xiao, and Lin]{chen2023control}
Weifeng Chen, Yatai Ji, Jie Wu, Hefeng Wu, Pan Xie, Jiashi Li, Xin Xia, Xuefeng Xiao, and Liang Lin.
\newblock Control-a-video: Controllable text-to-video generation with diffusion models.
\newblock \emph{arXiv preprint arXiv:2305.13840}, 2023{\natexlab{d}}.

\bibitem[Ding et~al.(2023)Ding, Ma, Dong, Zhang, Huang, Wang, Zheng, and Wei]{ding2023longnet}
Jiayu Ding, Shuming Ma, Li Dong, Xingxing Zhang, Shaohan Huang, Wenhui Wang, Nanning Zheng, and Furu Wei.
\newblock Longnet: Scaling transformers to 1,000,000,000 tokens 2023.
\newblock \emph{arXiv preprint arXiv:2307.02486}, 2023.

\bibitem[Farneb{\"a}ck(2003)]{farneback2003two}
Gunnar Farneb{\"a}ck.
\newblock Two-frame motion estimation based on polynomial expansion.
\newblock In \emph{Proceedings of the Scandinavian Conference on Image Analysis (SCIA)}, pages 363--370. Springer, 2003.

\bibitem[Geyer et~al.(2023)Geyer, Bar-Tal, Bagon, and Dekel]{geyer2023tokenflow}
Michal Geyer, Omer Bar-Tal, Shai Bagon, and Tali Dekel.
\newblock Tokenflow: Consistent diffusion features for consistent video editing.
\newblock \emph{arXiv preprint arXiv:2307.10373}, 2023.

\bibitem[Guo et~al.(2024)Guo, Zhang, Liu, Zhong, Zhang, Wan, and Zhang]{guo2024liveportrait}
Jianzhu Guo, Dingyun Zhang, Xiaoqiang Liu, Zhizhou Zhong, Yuan Zhang, Pengfei Wan, and Di Zhang.
\newblock Liveportrait: Efficient portrait animation with stitching and retargeting control.
\newblock \emph{arXiv preprint arXiv:2407.03168}, 2024.

\bibitem[Guo et~al.(2023{\natexlab{a}})Guo, Zheng, Hou, Gao, Deng, Ma, Hu, Zha, Huang, Wan, et~al.]{guo2023i2v}
Xun Guo, Mingwu Zheng, Liang Hou, Yuan Gao, Yufan Deng, Chongyang Ma, Weiming Hu, Zhengjun Zha, Haibin Huang, Pengfei Wan, et~al.
\newblock I2v-adapter: A general image-to-video adapter for video diffusion models.
\newblock \emph{arXiv preprint arXiv:2312.16693}, 2023{\natexlab{a}}.

\bibitem[Guo et~al.(2023{\natexlab{b}})Guo, Yang, Rao, Wang, Qiao, Lin, and Dai]{guo2023animatediff}
Yuwei Guo, Ceyuan Yang, Anyi Rao, Yaohui Wang, Yu Qiao, Dahua Lin, and Bo Dai.
\newblock Animatediff: Animate your personalized text-to-image diffusion models without specific tuning.
\newblock \emph{arXiv preprint arXiv:2307.04725}, 2023{\natexlab{b}}.

\bibitem[He et~al.(2024{\natexlab{a}})He, Liang, Wang, Cai, Xu, Guo, Wen, and Chen]{he2024lucidfusion}
Hao He, Yixun Liang, Luozhou Wang, Yuanhao Cai, Xinli Xu, Hao-Xiang Guo, Xiang Wen, and Yingcong Chen.
\newblock Lucidfusion: Generating 3d gaussians with arbitrary unposed images, 2024{\natexlab{a}}.

\bibitem[He et~al.(2024{\natexlab{b}})He, Xu, Guo, Wetzstein, Dai, Li, and Yang]{he2024cameractrl}
Hao He, Yinghao Xu, Yuwei Guo, Gordon Wetzstein, Bo Dai, Hongsheng Li, and Ceyuan Yang.
\newblock Cameractrl: Enabling camera control for text-to-video generation.
\newblock \emph{arXiv preprint arXiv:2404.02101}, 2024{\natexlab{b}}.

\bibitem[He et~al.(2024{\natexlab{c}})He, Li, Yin, Liang, Li, Zhou, Liu, Liu, and Chen]{he2024lotus}
Jing He, Haodong Li, Wei Yin, Yixun Liang, Leheng Li, Kaiqiang Zhou, Hongbo Liu, Bingbing Liu, and Ying-Cong Chen.
\newblock Lotus: Diffusion-based visual foundation model for high-quality dense prediction.
\newblock \emph{arXiv preprint arXiv:2409.18124}, 2024{\natexlab{c}}.

\bibitem[He et~al.(2022)He, Yang, Zhang, Shan, and Chen]{he2022latent}
Yingqing He, Tianyu Yang, Yong Zhang, Ying Shan, and Qifeng Chen.
\newblock Latent video diffusion models for high-fidelity video generation with arbitrary lengths.
\newblock \emph{arXiv preprint arXiv:2211.13221}, 2022.

\bibitem[Ho et~al.(2020)Ho, Jain, and Abbeel]{ho2020denoising}
Jonathan Ho, Ajay Jain, and Pieter Abbeel.
\newblock Denoising diffusion probabilistic models.
\newblock \emph{Advances in neural information processing systems}, 33:\penalty0 6840--6851, 2020.

\bibitem[Horn and Schunck(1981)]{horn1981determining}
Berthold~KP Horn and Brian~G Schunck.
\newblock Determining optical flow.
\newblock \emph{Artificial intelligence}, 17\penalty0 (1-3):\penalty0 185--203, 1981.

\bibitem[Hu et~al.(2021)Hu, Shen, Wallis, Allen-Zhu, Li, Wang, Wang, and Chen]{hu2021lora}
Edward~J Hu, Yelong Shen, Phillip Wallis, Zeyuan Allen-Zhu, Yuanzhi Li, Shean Wang, Lu Wang, and Weizhu Chen.
\newblock Lora: Low-rank adaptation of large language models.
\newblock \emph{arXiv preprint arXiv:2106.09685}, 2021.

\bibitem[Jeong et~al.(2024)Jeong, Chang, Park, and Ye]{jeong2024dreammotion}
Hyeonho Jeong, Jinho Chang, Geon~Yeong Park, and Jong~Chul Ye.
\newblock Dreammotion: Space-time self-similarity score distillation for zero-shot video editing.
\newblock \emph{arXiv preprint arXiv:2403.12002}, 2024.

\bibitem[Ke et~al.(2024)Ke, Obukhov, Huang, Metzger, Daudt, and Schindler]{ke2024repurposing}
Bingxin Ke, Anton Obukhov, Shengyu Huang, Nando Metzger, Rodrigo~Caye Daudt, and Konrad Schindler.
\newblock Repurposing diffusion-based image generators for monocular depth estimation.
\newblock In \emph{Proceedings of the IEEE/CVF Conference on Computer Vision and Pattern Recognition}, pages 9492--9502, 2024.

\bibitem[Lab and etc.(2024)]{opensoraplan}
PKU-Yuan Lab and Tuzhan~AI etc.
\newblock Open-sora-plan, 2024.

\bibitem[Li et~al.(2024)Li, Jain, and Shi]{li2024matting}
Jiachen Li, Jitesh Jain, and Humphrey Shi.
\newblock Matting anything.
\newblock In \emph{Proceedings of the IEEE/CVF Conference on Computer Vision and Pattern Recognition}, pages 1775--1785, 2024.

\bibitem[Lin et~al.(2023)Lin, Gao, Huang, Kim, Wang, Zwicker, and Saraf]{lin2023omnimatterf}
Geng Lin, Chen Gao, Jia-Bin Huang, Changil Kim, Yipeng Wang, Matthias Zwicker, and Ayush Saraf.
\newblock Omnimatterf: Robust omnimatte with 3d background modeling.
\newblock In \emph{Proceedings of the IEEE/CVF International Conference on Computer Vision}, pages 23471--23480, 2023.

\bibitem[Lin et~al.(2021)Lin, Ryabtsev, Sengupta, Curless, Seitz, and Kemelmacher-Shlizerman]{lin2021real}
Shanchuan Lin, Andrey Ryabtsev, Soumyadip Sengupta, Brian~L Curless, Steven~M Seitz, and Ira Kemelmacher-Shlizerman.
\newblock Real-time high-resolution background matting.
\newblock In \emph{Proceedings of the IEEE/CVF Conference on Computer Vision and Pattern Recognition}, pages 8762--8771, 2021.

\bibitem[Lin et~al.(2022)Lin, Yang, Saleemi, and Sengupta]{lin2022robust}
Shanchuan Lin, Linjie Yang, Imran Saleemi, and Soumyadip Sengupta.
\newblock Robust high-resolution video matting with temporal guidance.
\newblock In \emph{Proceedings of the IEEE/CVF Winter Conference on Applications of Computer Vision}, pages 238--247, 2022.

\bibitem[Ling et~al.(2024)Ling, Bu, Zhang, Dong, Zang, Wu, Chen, Wang, and Jin]{ling2024motionclone}
Pengyang Ling, Jiazi Bu, Pan Zhang, Xiaoyi Dong, Yuhang Zang, Tong Wu, Huaian Chen, Jiaqi Wang, and Yi Jin.
\newblock Motionclone: Training-free motion cloning for controllable video generation.
\newblock \emph{arXiv preprint arXiv:2406.05338}, 2024.

\bibitem[Liu et~al.(2024)Liu, Zhang, Li, Lin, and Jia]{liu2024video}
Shaoteng Liu, Yuechen Zhang, Wenbo Li, Zhe Lin, and Jiaya Jia.
\newblock Video-p2p: Video editing with cross-attention control.
\newblock In \emph{Proceedings of the IEEE/CVF Conference on Computer Vision and Pattern Recognition}, pages 8599--8608, 2024.

\bibitem[Liu et~al.(2022)Liu, Gong, and Liu]{liu2022flow}
Xingchao Liu, Chengyue Gong, and Qiang Liu.
\newblock Flow straight and fast: Learning to generate and transfer data with rectified flow.
\newblock \emph{arXiv preprint arXiv:2209.03003}, 2022.

\bibitem[Long et~al.(2024)Long, Guo, Lin, Liu, Dou, Liu, Ma, Zhang, Habermann, Theobalt, et~al.]{long2024wonder3d}
Xiaoxiao Long, Yuan-Chen Guo, Cheng Lin, Yuan Liu, Zhiyang Dou, Lingjie Liu, Yuexin Ma, Song-Hai Zhang, Marc Habermann, Christian Theobalt, et~al.
\newblock Wonder3d: Single image to 3d using cross-domain diffusion.
\newblock In \emph{Proceedings of the IEEE/CVF Conference on Computer Vision and Pattern Recognition}, pages 9970--9980, 2024.

\bibitem[Luo et~al.(2024)Luo, Ceylan, Yoon, Zhao, Philip, Fr{\"u}hst{\"u}ck, Li, Richardt, and Wang]{luo2024intrinsicdiffusion}
Jundan Luo, Duygu Ceylan, Jae~Shin Yoon, Nanxuan Zhao, Julien Philip, Anna Fr{\"u}hst{\"u}ck, Wenbin Li, Christian Richardt, and Tuanfeng Wang.
\newblock Intrinsicdiffusion: Joint intrinsic layers from latent diffusion models.
\newblock In \emph{ACM SIGGRAPH 2024 Conference Papers}, pages 1--11, 2024.

\bibitem[Ma et~al.(2023)Ma, Lewis, and Kleijn]{ma2023trailblazer}
Wan-Duo~Kurt Ma, John~P Lewis, and W~Bastiaan Kleijn.
\newblock Trailblazer: Trajectory control for diffusion-based video generation.
\newblock \emph{arXiv preprint arXiv:2401.00896}, 2023.

\bibitem[Ma et~al.(2024)Ma, Wang, Jia, Chen, Liu, Li, Chen, and Qiao]{ma2024latte}
Xin Ma, Yaohui Wang, Gengyun Jia, Xinyuan Chen, Ziwei Liu, Yuan-Fang Li, Cunjian Chen, and Yu Qiao.
\newblock Latte: Latent diffusion transformer for video generation.
\newblock \emph{arXiv preprint arXiv:2401.03048}, 2024.

\bibitem[Niu et~al.(2024)Niu, Cun, Wang, Zhang, Shan, and Zheng]{niu2024mofa}
Muyao Niu, Xiaodong Cun, Xintao Wang, Yong Zhang, Ying Shan, and Yinqiang Zheng.
\newblock Mofa-video: Controllable image animation via generative motion field adaptions in frozen image-to-video diffusion model.
\newblock \emph{arXiv preprint arXiv:2405.20222}, 2024.

\bibitem[Qi et~al.(2023)Qi, Cun, Zhang, Lei, Wang, Shan, and Chen]{qi2023fatezero}
Chenyang Qi, Xiaodong Cun, Yong Zhang, Chenyang Lei, Xintao Wang, Ying Shan, and Qifeng Chen.
\newblock Fatezero: Fusing attentions for zero-shot text-based video editing.
\newblock In \emph{Proceedings of the IEEE/CVF International Conference on Computer Vision}, pages 15932--15942, 2023.

\bibitem[Qin et~al.(2023)Qin, Ke, Ma, Danelljan, Tai, Tang, Liu, and Yu]{qin2023bimatting}
Haotong Qin, Lei Ke, Xudong Ma, Martin Danelljan, Yu-Wing Tai, Chi-Keung Tang, Xianglong Liu, and Fisher Yu.
\newblock Bimatting: Efficient video matting via binarization.
\newblock \emph{Advances in Neural Information Processing Systems}, 36:\penalty0 43307--43321, 2023.

\bibitem[Ravi et~al.(2024)Ravi, Gabeur, Hu, Hu, Ryali, Ma, Khedr, R{\"a}dle, Rolland, Gustafson, Mintun, Pan, Alwala, Carion, Wu, Girshick, Doll{\'a}r, and Feichtenhofer]{ravi2024sam2}
Nikhila Ravi, Valentin Gabeur, Yuan-Ting Hu, Ronghang Hu, Chaitanya Ryali, Tengyu Ma, Haitham Khedr, Roman R{\"a}dle, Chloe Rolland, Laura Gustafson, Eric Mintun, Junting Pan, Kalyan~Vasudev Alwala, Nicolas Carion, Chao-Yuan Wu, Ross Girshick, Piotr Doll{\'a}r, and Christoph Feichtenhofer.
\newblock Sam 2: Segment anything in images and videos.
\newblock \emph{arXiv preprint arXiv:2408.00714}, 2024.

\bibitem[Rombach et~al.(2022)Rombach, Blattmann, Lorenz, Esser, and Ommer]{rombach2022high}
Robin Rombach, Andreas Blattmann, Dominik Lorenz, Patrick Esser, and Bj{\"o}rn Ommer.
\newblock High-resolution image synthesis with latent diffusion models.
\newblock In \emph{Proceedings of the IEEE/CVF conference on computer vision and pattern recognition}, pages 10684--10695, 2022.

\bibitem[Su et~al.(2024)Su, Ahmed, Lu, Pan, Bo, and Liu]{su2024roformer}
Jianlin Su, Murtadha Ahmed, Yu Lu, Shengfeng Pan, Wen Bo, and Yunfeng Liu.
\newblock Roformer: Enhanced transformer with rotary position embedding.
\newblock \emph{Neurocomputing}, 568:\penalty0 127063, 2024.

\bibitem[Team(2024)]{genmo2024mochi}
Genmo Team.
\newblock Mochi, 2024.

\bibitem[Unterthiner et~al.(2019)Unterthiner, van Steenkiste, Kurach, Marinier, Michalski, and Gelly]{unterthiner2019fvd}
Thomas Unterthiner, Sjoerd van Steenkiste, Karol Kurach, Rapha{\"e}l Marinier, Marcin Michalski, and Sylvain Gelly.
\newblock Fvd: A new metric for video generation.
\newblock 2019.

\bibitem[Vainer et~al.(2024)Vainer, Boss, Parger, Kutsy, De~Nigris, Rowles, Perony, and Donn{\'e}]{vainer2024collaborative}
Shimon Vainer, Mark Boss, Mathias Parger, Konstantin Kutsy, Dante De~Nigris, Ciara Rowles, Nicolas Perony, and Simon Donn{\'e}.
\newblock Collaborative control for geometry-conditioned pbr image generation.
\newblock \emph{arXiv preprint arXiv:2402.05919}, 2024.

\bibitem[Wang et~al.(2023)Wang, Yuan, Chen, Zhang, Wang, and Zhang]{wang2023modelscope}
Jiuniu Wang, Hangjie Yuan, Dayou Chen, Yingya Zhang, Xiang Wang, and Shiwei Zhang.
\newblock Modelscope text-to-video technical report.
\newblock \emph{arXiv preprint arXiv:2308.06571}, 2023.

\bibitem[Wang et~al.(2024{\natexlab{a}})Wang, Shen, Liang, Tao, Wan, Zhang, Li, and Chen]{wang2024motion}
Luozhou Wang, Guibao Shen, Yixun Liang, Xin Tao, Pengfei Wan, Di Zhang, Yijun Li, and Yingcong Chen.
\newblock Motion inversion for video customization.
\newblock \emph{arXiv preprint arXiv:2403.20193}, 2024{\natexlab{a}}.

\bibitem[Wang et~al.(2020)Wang, Li, Khabsa, Fang, and Ma]{wang2020linformer}
Sinong Wang, Belinda~Z Li, Madian Khabsa, Han Fang, and Hao Ma.
\newblock Linformer: Self-attention with linear complexity.
\newblock \emph{arXiv preprint arXiv:2006.04768}, 2020.

\bibitem[Wang et~al.(2024{\natexlab{b}})Wang, Yuan, Zhang, Chen, Wang, Zhang, Shen, Zhao, and Zhou]{wang2024videocomposer}
Xiang Wang, Hangjie Yuan, Shiwei Zhang, Dayou Chen, Jiuniu Wang, Yingya Zhang, Yujun Shen, Deli Zhao, and Jingren Zhou.
\newblock Videocomposer: Compositional video synthesis with motion controllability.
\newblock \emph{Advances in Neural Information Processing Systems}, 36, 2024{\natexlab{b}}.

\bibitem[Wang et~al.(2024{\natexlab{c}})Wang, Li, Wang, Liu, Gu, Chuang, and Satoh]{wang2024matting}
Zhixiang Wang, Baiang Li, Jian Wang, Yu-Lun Liu, Jinwei Gu, Yung-Yu Chuang, and Shin'Ichi Satoh.
\newblock Matting by generation.
\newblock In \emph{ACM SIGGRAPH 2024 Conference Papers}, pages 1--11, 2024{\natexlab{c}}.

\bibitem[Wang et~al.(2024{\natexlab{d}})Wang, Yuan, Wang, Li, Chen, Xia, Luo, and Shan]{wang2024motionctrl}
Zhouxia Wang, Ziyang Yuan, Xintao Wang, Yaowei Li, Tianshui Chen, Menghan Xia, Ping Luo, and Ying Shan.
\newblock Motionctrl: A unified and flexible motion controller for video generation.
\newblock In \emph{ACM SIGGRAPH 2024 Conference Papers}, pages 1--11, 2024{\natexlab{d}}.

\bibitem[Wu et~al.(2023)Wu, Ge, Wang, Lei, Gu, Shi, Hsu, Shan, Qie, and Shou]{wu2023tune}
Jay~Zhangjie Wu, Yixiao Ge, Xintao Wang, Stan~Weixian Lei, Yuchao Gu, Yufei Shi, Wynne Hsu, Ying Shan, Xiaohu Qie, and Mike~Zheng Shou.
\newblock Tune-a-video: One-shot tuning of image diffusion models for text-to-video generation.
\newblock In \emph{Proceedings of the IEEE/CVF International Conference on Computer Vision}, pages 7623--7633, 2023.

\bibitem[Yang et~al.(2024{\natexlab{a}})Yang, Kang, Huang, Xu, Feng, and Zhao]{yang2024depth}
Lihe Yang, Bingyi Kang, Zilong Huang, Xiaogang Xu, Jiashi Feng, and Hengshuang Zhao.
\newblock Depth anything: Unleashing the power of large-scale unlabeled data.
\newblock In \emph{Proceedings of the IEEE/CVF Conference on Computer Vision and Pattern Recognition}, pages 10371--10381, 2024{\natexlab{a}}.

\bibitem[Yang et~al.(2023{\natexlab{a}})Yang, Chen, Chen, Fang, Liu, and Chen]{yang2023defect}
Shuai Yang, Zhifei Chen, Pengguang Chen, Xi Fang, Shu Liu, and Yingcong Chen.
\newblock Defect spectrum: A granular look of large-scale defect datasets with rich semantics, 2023{\natexlab{a}}.

\bibitem[Yang et~al.(2023{\natexlab{b}})Yang, Zhou, Liu, and Loy]{yang2023rerender}
Shuai Yang, Yifan Zhou, Ziwei Liu, and Chen~Change Loy.
\newblock Rerender a video: Zero-shot text-guided video-to-video translation.
\newblock In \emph{SIGGRAPH Asia 2023 Conference Papers}, pages 1--11, 2023{\natexlab{b}}.

\bibitem[Yang et~al.(2024{\natexlab{b}})Yang, Teng, Zheng, Ding, Huang, Xu, Yang, Hong, Zhang, Feng, et~al.]{yang2024cogvideox}
Zhuoyi Yang, Jiayan Teng, Wendi Zheng, Ming Ding, Shiyu Huang, Jiazheng Xu, Yuanming Yang, Wenyi Hong, Xiaohan Zhang, Guanyu Feng, et~al.
\newblock Cogvideox: Text-to-video diffusion models with an expert transformer.
\newblock \emph{arXiv preprint arXiv:2408.06072}, 2024{\natexlab{b}}.

\bibitem[Yao et~al.(2024{\natexlab{a}})Yao, Wang, Yang, and Wang]{yao2024vitmatte}
Jingfeng Yao, Xinggang Wang, Shusheng Yang, and Baoyuan Wang.
\newblock Vitmatte: Boosting image matting with pre-trained plain vision transformers.
\newblock \emph{Information Fusion}, 103:\penalty0 102091, 2024{\natexlab{a}}.

\bibitem[Yao et~al.(2024{\natexlab{b}})Yao, Wang, Ye, and Liu]{yao2024matte}
Jingfeng Yao, Xinggang Wang, Lang Ye, and Wenyu Liu.
\newblock Matte anything: Interactive natural image matting with segment anything model.
\newblock \emph{Image and Vision Computing}, 147:\penalty0 105067, 2024{\natexlab{b}}.

\bibitem[Yin et~al.(2023)Yin, Wu, Liang, Shi, Li, Ming, and Duan]{yin2023dragnuwa}
Shengming Yin, Chenfei Wu, Jian Liang, Jie Shi, Houqiang Li, Gong Ming, and Nan Duan.
\newblock Dragnuwa: Fine-grained control in video generation by integrating text, image, and trajectory.
\newblock \emph{arXiv preprint arXiv:2308.08089}, 2023.

\bibitem[Zeng et~al.(2024)Zeng, Deschaintre, Georgiev, Hold-Geoffroy, Hu, Luan, Yan, and Ha{\v{s}}an]{zeng2024rgb}
Zheng Zeng, Valentin Deschaintre, Iliyan Georgiev, Yannick Hold-Geoffroy, Yiwei Hu, Fujun Luan, Ling-Qi Yan, and Milo{\v{s}} Ha{\v{s}}an.
\newblock Rgb\(\leftrightarrow\)x: Image decomposition and synthesis using material-and lighting-aware diffusion models.
\newblock In \emph{ACM SIGGRAPH 2024 Conference Papers}, pages 1--11, 2024.

\bibitem[Zhang et~al.(2023)Zhang, Wu, Liu, Zhao, Ran, Gu, Gao, and Shou]{zhang2023show}
David~Junhao Zhang, Jay~Zhangjie Wu, Jia-Wei Liu, Rui Zhao, Lingmin Ran, Yuchao Gu, Difei Gao, and Mike~Zheng Shou.
\newblock Show-1: Marrying pixel and latent diffusion models for text-to-video generation.
\newblock \emph{arXiv preprint arXiv:2309.15818}, 2023.

\bibitem[Zhang et~al.(2024)Zhang, Li, Le, Shou, Xiong, and Sahoo]{zhang2024moonshot}
David~Junhao Zhang, Dongxu Li, Hung Le, Mike~Zheng Shou, Caiming Xiong, and Doyen Sahoo.
\newblock Moonshot: Towards controllable video generation and editing with multimodal conditions.
\newblock \emph{arXiv preprint arXiv:2401.01827}, 2024.

\bibitem[Zhang and Agrawala(2024)]{zhang2024transparent}
Lvmin Zhang and Maneesh Agrawala.
\newblock Transparent image layer diffusion using latent transparency.
\newblock \emph{arXiv preprint arXiv:2402.17113}, 2024.

\bibitem[Zheng et~al.(2024)Zheng, Peng, Yang, Shen, Li, Liu, Zhou, Li, and You]{opensora}
Zangwei Zheng, Xiangyu Peng, Tianji Yang, Chenhui Shen, Shenggui Li, Hongxin Liu, Yukun Zhou, Tianyi Li, and Yang You.
\newblock Open-sora: Democratizing efficient video production for all, 2024.

\bibitem[Zhu et~al.(2021)Zhu, Ping, Xiao, Shoeybi, Goldstein, Anandkumar, and Catanzaro]{zhu2021long}
Chen Zhu, Wei Ping, Chaowei Xiao, Mohammad Shoeybi, Tom Goldstein, Anima Anandkumar, and Bryan Catanzaro.
\newblock Long-short transformer: Efficient transformers for language and vision.
\newblock \emph{Advances in neural information processing systems}, 34:\penalty0 17723--17736, 2021.

\end{thebibliography}
}

\end{document}